\documentclass[conference]{IEEEtran}
\IEEEoverridecommandlockouts
\usepackage{cite}
\usepackage{amsmath,amssymb,amsfonts}
\usepackage{algorithmic}
\usepackage{graphicx}
\usepackage{textcomp}
\usepackage{xcolor}
\usepackage{hyperref}
\usepackage{bm}
\usepackage{amssymb}% http://ctan.org/pkg/amssymb
\usepackage{amsfonts}
\usepackage{pifont}% http://ctan.org/pkg/pifont

\usepackage{textcomp}
\usepackage{booktabs} 
\usepackage{adjustbox}
\usepackage{graphicx}

\usepackage{caption}
\usepackage{subcaption}
\usepackage{float}

\usepackage{longtable}
\usepackage{tabularx}
\usepackage{environ}
\usepackage{multirow}
\usepackage{color, colortbl} % Colored background for table
\usepackage{booktabs}
\usepackage{xcolor}

\usepackage{makecell}
\usepackage[utf8]{inputenc} % allow utf-8 input
\usepackage[T1]{fontenc}    % use 8-bit T1 fonts
\usepackage{textcomp}
\usepackage{adjustbox}

\usepackage{array}
\usepackage{paralist}
\usepackage{rotating}
\usepackage{pgfplotstable}
\usepackage{pgfplots}

\usepgfplotslibrary{fillbetween}
\usepackage{algorithmic}
\usepackage[linesnumbered,ruled]{algorithm2e}
\usepackage{pifont}
\newcommand{\cm}{{\color{Green1}{\ding{51}}}}
\newcommand{\xm}{{\color{Red1}{\ding{55}}}}
\usepackage{amsmath}
\usepackage{amssymb}

\usepackage{siunitx}

\pgfplotsset{compat=1.7}

\usepgfplotslibrary{groupplots}
\definecolor{Blue1}{RGB}{9, 0, 255}
\definecolor{Red1}{RGB}{204, 0, 0}
\definecolor{Green1}{RGB}{0, 92, 78}
\definecolor{DarkGreen}{RGB}{85,168,104}

\definecolor{entropy}{named}{black}
\definecolor{not_normalized}{named}{orange}
\definecolor{normalized}{named}{green}

\definecolor{clr_random}{RGB}{196,78,82}  % Red

\definecolor{clr_entropy}{RGB}{85,168,104}  % Green

\definecolor{clr_true_loss}{RGB}{76,114,176}  % Blue

\definecolor{clr_coverage}{RGB}{221,132,82}  % Orange

\definecolor{clr_lp_dense}{RGB}{218,139,195}  % pink

\definecolor{clr_r20}{RGB}{196,78,82}  % Red

\definecolor{clr_r18}{RGB}{85,168,104}  % Green

\definecolor{clr_r50}{RGB}{76,114,176}  % Blue

\definecolor{clr_xception}{RGB}{221,132,82}  % Orange

\definecolor{Green-1}{rgb}{0.0, 0.5, 0.0}
\definecolor{Green-2}{rgb}{0.55, 0.71, 0.0}
\definecolor{Yellow-1}{rgb}{1.0, 0.75, 0.0}
\definecolor{Orange-1}{rgb}{1.0, 0.49, 0.0}
\definecolor{Red-1}{rgb}{0.9, 0.0, 0.13}
\usepackage{makecell}
\usepackage{tikz}
\usepackage{pgfplots, pgfplotstable}

\makeatletter % changes the catcode of @ to 11
\newcommand{\linebreakand}{%
  \end{@IEEEauthorhalign}
  \hfill\mbox{}\par
  \mbox{}\hfill\begin{@IEEEauthorhalign}
}
\makeatother % changes the catcode of @ back to 12

\def\BibTeX{{\rm B\kern-.05em{\sc i\kern-.025em b}\kern-.08em
    T\kern-.1667em\lower.7ex\hbox{E}\kern-.125emX}}
\begin{document}

% Behzad removed ... for *Robust* HW-aware ... to save a few lines which helped a lot
\title{HiFi-LLP: High-Fidelity, Low-Cost Latency Predictors with Confidence for Robust HW-NAS
\thanks{\copyright~2025 IEEE. Personal use of this material is permitted. Permission from IEEE must be obtained for all other uses, in any current or future media, including reprinting/republishing this material for advertising or promotional purposes, creating new collective works, for resale or redistribution to servers or lists, or reuse of any copyrighted component of this work in other works. Published in the Proceedings of the 2025 IEEE 38th International System-on-Chip Conference (SOCC). DOI: \texttt{10.1109/SOCC66126.2025.11235466}}
% \\
}

\author{
Shambhavi Balamuthu Sampath$^{1,2\dagger}$
~~ Behzad Shomali$^{3\dagger}$
~~ Nael Fasfous$^{2}$
~~ Moritz Thoma$^{1,2}$\\
~~ Judeson Anthony Fernando$^{1}$
~~Lukas Frickenstein$^{2}$
~~Pierpaolo Mori$^{2}$
~~ Manoj Rohit Vemparala$^{2}$\\
~~ Alexander Frickenstein$^{2}$
~~ Walter Stechele$^{1}$\vspace{+1ex}\\
\normalsize $\dagger$ Equal Contribution ~~
\normalsize $^1$Technical University of Munich ~~ $^2$BMW Group ~~ $^3$University of Bonn

\vspace{-0.64ex}
}

% \author{
% \IEEEauthorblockN{Shambhavi Balamuthu Sampath\IEEEauthorrefmark{1}}
% \IEEEauthorblockA{
%  \textit{BMW AG}\\
% % Munich, Germany \\
% shambhavi.balamuthu-sampath@bmw.de}
% \and
% \IEEEauthorblockN{Behzad Shomali\IEEEauthorrefmark{1}}
% \IEEEcompsocitemizethanks{\IEEEcompsocthanksitem\IEEEauthorrefmark{1}equal contribution}
% \IEEEauthorblockA{
%  \textit{University of Bonn}\\
% % Bonn, Germany\\
% behzad.shomali@uni-bonn.de}
% \and
% \IEEEauthorblockN{Nael Fasfous}
% \IEEEauthorblockA{
% \textit{BMW AG}\\
% % Munich, Germany \\
% nael.fasfous@bmw.de}
% \and
% \IEEEauthorblockN{Moritz Thoma}
% \IEEEauthorblockA{
% \textit{BMW AG}\\
% % Munich, Germany \\
% moritz.thoma@bmw.de}
% \and
% \IEEEauthorblockN{Judeson Anthony Fernando}
% \IEEEauthorblockA{
% \textit{Technical University of Munich}\\
% % Munich, Germany \\
% judeson.anthony-fernando@tum.de}
% \and
% \IEEEauthorblockN{Lukas Frickenstein}
% \IEEEauthorblockA{
% \textit{BMW AG}\\
% % Munich, Germany \\
% lukas.frickenstein@bmw.de}
% \and
% \IEEEauthorblockN{Pierpaolo Mori}
% \IEEEauthorblockA{
% \textit{‌BMW AG}\\
% % Munich, Germany \\
% pierpaolo.mori@bmw.de}
% \and
% \IEEEauthorblockN{Manoj Rohit Vemparala}
% \IEEEauthorblockA{
% \textit{‌BMW AG}\\
% % Munich, Germany \\
% manoj-rohit.vemparala@bmw.de}
% \and
% \linebreakand
% \IEEEauthorblockN{Alexander Frickenstein}
% \IEEEauthorblockA{
% \textit{‌BMW AG}\\
% % Munich, Germany \\
% alexander.frickenstein@bmw.de}
% \and
% \IEEEauthorblockN{Walter Stechele}
% \IEEEauthorblockA{
% \textit{‌Technical University of Munich}\\
% % Munich, Germany \\
% walter.stechele@tum.de}
% }

\maketitle

\begin{abstract}
With deep neural networks (DNNs) increasingly deployed on edge devices, hardware (HW)-aware optimization techniques—such as HW-aware compression and HW-aware neural architecture search (HW-NAS)—have become essential. These methods rely on real feedback from the target hardware to tailor DNN architectures for efficient deployment. While the search can be parallelized, latency measurements via hardware-in-the-loop (HIL) remain a bottleneck due to their sequential nature. Recent approaches use latency predictors to replace costly HIL feedback, but challenges persist: (1) platform-specific predictors often require tens of thousands of samples, and (2) inaccurate predictions can mislead the NAS process. To address this, we introduce HiFi-LLP, a high-fidelity, low-cost latency predictor based on graph attention networks, augmented with a confidence metric. HiFi-LLP outperforms prior platform-specific predictors by up to 9 percentage points (p.p.) in the 10\% accuracy bound and achieves a Spearman's rank correlation of up to 0.996 across six devices in the LatBench dataset. We further propose a hybrid NAS framework that routes low-confidence predictions to HIL, achieving up to 8.6× speedup compared to typical NAS while maintaining a competitive Pareto front. 
% Code is available at \href{https://github.com/shamvbs/HiFi-LLP}{https://github.com/shamvbs/HiFi-LLP}. 
\footnote{\small Correspondence to: \href{mailto:shambhavi.balamuthu-sampath@bmw.de}{shambhavi.balamuthu-sampath@tum.de}} 
\end{abstract}

% With this, we position HiFi-LLP as a sample-efficient, high fidelity latency predictor that provides confidence scores for a reliable and robust HW-NAS. 

% Structural reparametrization (SRP) involves a bipartite network design, (1) an expanded multi-branch train-time network that is equivalent to a (2) collapsed lightweight inference-time network. This means, at the cost of a longer training time of the expanded network, the task metrics of the efficient inference-time network are improved. Recent works have adopted SRP for handcrafting efficient deep neural networks and have proposed universal multi-branch SRP blocks to expand only the convolutional (CONV) layers during training. Furthermore, the training schemes involve uniformly expanding \textit{every} CONV layer of a given DNN, which becomes impractical for modern, large DNNs. 


\section{Introduction}
Safety-critical domains, such as autonomous driving, robotics, and industrial control, require algorithms that ensure high performance at real-time speeds. As DNNs are increasingly applied in these areas, the focus has shifted from merely achieving high accuracy to balancing accuracy with low latency. 
% This shift demands the customization of architectures to meet both objectives for specific target devices, turning the task into a complex search problem within a vast design space. To address this, automated NAS solutions have evolved from optimizing for accuracy to HW-aware NAS, which takes into account both accuracy and hardware constraints, such as memory footprint and latency.
This shift requires architecture customization for specific target devices to meet both objectives, turning the task into a complex search across a large design space. To address this, NAS has evolved from optimizing for accuracy to HW-aware NAS (HW-NAS), which considers both accuracy and hardware constraints such as memory footprint and latency. % save ~2 lines -- Accepted!

% convert HW-aware NAS --> HW-NAS across the manuscript
Conventional HW-NAS frameworks, including HW-aware compression search, often rely on proxy metrics like floating point operations per second (FLOPS) or build exhaustive look-up tables (LUTs) for the search space. The latter is accurate but is infeasible for vast search spaces, while the former is inaccurate, as lower FLOPS do not necessarily translate to lower latencies \cite{akhauri2024latency}. Therefore, recent works, especially for rapid NAS, have increasingly adopted latency predictors \cite{shala2023transfer, akhauri2024latency}. Developed as a software solution, latency predictors are parallelizable with compute and eliminate the bottleneck of slow and sequential HIL measurements. State-of-the-art (SOTA) latency predictors often encounter a trade-off between latency predictor building time and their prediction accuracy. It is intuitive that with large amounts of training data resulting from exhaustive measurements, one can build an accurate latency predictor for NAS. However, latency is influenced by changes in compiler optimizations and hardware updates. As a result, the reliability of a latency predictor becomes questionable with each periodic update. To address this, researchers used meta-learning to train latency predictors on a group of devices, allowing them to adapt more easily to similar, unseen devices. While this approach involves a one-time data collection and training overhead, these predictors do not perform well on dissimilar, unseen devices. Additionally, because they are trained on multiple devices, they require specific hardware descriptors to differentiate between them \cite{akhauri2023multi, lee2021help}. In contrast, our work introduces a platform-specific, low-cost, and high-fidelity latency predictor that offers reliable predictions along with an actual certainty metric.
Our contributions can be summarized as follows:
\begin{itemize}
    \item We introduce HiFi-LLP, a high-fidelity, low-cost latency predictor, that can be trained from scratch with just 100 samples, providing predictions with their confidence.

    % \item We demonstrate that HiFi-LLP outperforms SOTA platform-specific latency predictors across all devices in the LatBench dataset, as measured by Spearman's rank correlation coefficient and the 10\% accuracy metric. 

    \item HiFi-LLP outperforms SOTA platform-specific latency predictors across all LatBench \cite{dudziak2020brp} devices as measured by Spearman’s rank and the 10\% accuracy metric. % saved one line! Ok.
    
    \item We leverage the confidence metric of HiFi-LLP to propose a robust, \textit{hybrid} HW-NAS framework. 
    
    \item We finally show that Bayesian optimization (BO)-based data sampler surpasses other SOTA samplers in low-data scenarios for low-cost latency predictors.   
\end{itemize}

\section{Related Works}
%Latency predictors, though niche, are key for accelerating NAS. 
We categorize SOTA latency predictors into: \textit{platform-specific}, tailored to a single device, and \textit{quasi-generalized}, meta-trained on multiple devices to adapt to similar unseen hardware. Table \ref{tab:sota_works} outlines five key aspects: (1) \textit{no pretraining}, if no prior training on other devices/datasets is needed; (2) \textit{GNN-based}, if graph neural networks (GNNs) are used for end-to-end prediction; (3) \textit{sample-efficient}, if training requires 200 or less measurement samples; (4) \textit{guided sampling}, if smarter sampling strategies are used over random sampling; and (5) \textit{probabilistic}, if the predictor quantifies uncertainty.
%\bgroup
%\setlength{\tabcolsep}{2pt}
\begin{table}
    \centering
    \vspace{0.11in}
    \resizebox{\columnwidth}{!}{
        \begin{tabular}{l|ccc|cccc}
        \toprule \multicolumn{1}{c}{} & \multicolumn{3}{c}{\small\textbf{Quasi-generalized}} & \multicolumn{4}{c}{\small\textbf{Platform-specific}}\\
        \cmidrule(rl){2-4} \cmidrule(l){5-8}
            & \makecell{\textbf{\textit{HELP}}\cite{lee2021help}} & \makecell{\textbf{\textit{MultiPredict}} \cite{akhauri2023multi}} & \makecell{\textbf{\textit{NASFLAT}}  \cite{akhauri2024latency}} & \makecell{\textbf{\textit{nn-Meter}}  \cite{zhang2021nn}} & \makecell{\textbf{\textit{BRP-NAS}}  \cite{dudziak2020brp}}  & \makecell{\textbf{\textit{HiFi-SAGE}}  \cite{shambhavi25hifisage}} &  \textbf{\textit{Ours}} \\ \hline %\midrule 

            No pretraining & \xm & \xm & \xm & \xm & \cm & \cm & \cm\\
            GNN-based & \cm & \cm & \cm & \xm & \cm & \cm & \cm\\
            Sample-efficient & \cm & \cm & \cm & \xm & \xm & \cm & \cm\\
            Guided sampling & \xm & \cm & \cm & \cm & \xm & \xm & \cm\\
            Probabilistic & \xm & \xm & \xm & \xm & \xm & \xm & \cm\\
        \bottomrule
        \end{tabular}
     }
    \caption{Overview of the related works.}
    \vspace{-0.2in}
    \label{tab:sota_works}
\end{table}

%\egroup

HELP \cite{lee2021help} is meta-trained on a pool of 18 devices and claims to generalize to newer unseen devices with few-shot adaptation.  Multipredict \cite{akhauri2023multi} addresses the expensive meta-training by supplementing it with hardware-specific descriptors. However, the quality of these hardware descriptors largely influences the prediction quality. This becomes especially tricky with respect to certainty in the low-data regime. In contrast, we resort to simple platform-specific latency predictors, similar to HiFi-SAGE \cite{shambhavi25hifisage}, and explore its design space to propose an architecture for a latency predictor that can provide certainty in its predictions.
Among the SOTA platform-specific latency predictors, nn-Meter \cite{zhang2021nn} is the most accurate but nonetheless relies on collecting up to 26k samples. Their biggest drawback is that they assume sequential kernel execution, which is not representative of modern hardware.
BRP-NAS \cite{dudziak2020brp} was the first end-to-end graph convolutional network (GCN)-based predictor, which required 900 samples and introduced a binary relation module to improve accuracy.
% BRP-NAS \cite{dudziak2020brp} was the first to propose an end-to-end graph convolutional network (GCN)-based latency predictor. They propose rather sample-efficient training with 900 samples and propose a binary relation predictor that boosts accuracy prediction for NAS. 
% HiFi-SAGE \cite{shambhavi25hifisage} outperformed BRP-NAS via architectural improvements in the sample-efficient training regime. They encode DNNs on an operator level and leverage GraphSAGE layers with residual connections. 
HiFi-SAGE improves on BRP-NAS with architectural improvements, operator-level encoding, and GraphSAGE layers.
However, both BRP-NAS and HiFi-SAGE rely on random sampling and report averages over 100 runs, showing noticeable variance, which raises concerns about their reliability in real-world applications. This highlights a key research gap in the stability of sample-efficient, platform-specific predictors, motivating the need for exploring other sampling techniques and uncertainty estimation, particularly in low-data regimes.

\textbf{\textit{No pretraining}} predictors 
% that avoid pretraining
are advantageous in iterative, data-driven workflows with frequent hardware or compiler updates. Among the studied works, only BRP-NAS \cite{dudziak2020brp} and HiFi-SAGE \cite{shambhavi25hifisage} operate without extensive pretraining, while others require large datasets. Although meta-trained models adapt easily to similar devices, we argue that platform-specific predictors—when carefully designed—offer simpler, effective alternatives even in low-data settings. 
\textbf{\textit{GNN-based}} predictors can effectively capture the structural properties of DNNs, leading to more accurate latency predictions. Except for nn-meter \cite{zhang2021nn}, all the works in SOTA leverage GNNs. However, most works have not explored beyond simple graph convolutional layers for the latency predictor architecture. Differently, we exploit graph-attention transformers (GATs) in this work. 
\textbf{\textit{Sample-efficient}} % Such 
predictors are crucial to reducing the amount of training data required, thus eliminating bottlenecks in rapid NAS methods, as shown in \cite{lee2021help}. All the quasi-generalized predictors have a measurement-efficient adaptation phase to similar unseen devices. This works well where the underlying target is positively correlated with the training data. In contrast, the platform-specific nn-Meter \cite{zhang2021nn} requires 26K samples for training \cite{dudziak2020brp}. While BRP-NAS \cite{dudziak2020brp} requires 900 samples and HiFi-SAGE \cite{shambhavi25hifisage} only needs 100, both of them report their results as an average over 100 random seeds. Our experiments reveal a high variance between seed results, rendering these averages and performances unrepresentative of real-world scenarios.
\textbf{\textit{Guided sampling}} strategies 
% strategic sampling 
improve predictor accuracy and robustness over random sampling. MultiPredict \cite{akhauri2023multi} introduces a search space independent encoding, based on zero-cost proxies 
% that enables cross-task and cross-device predictor transfer,
but does not explicitly employ architecture-aware sampling in its core methodology. nn-Meter \cite{zhang2021nn} uses adaptive sampling, iteratively selecting the most informative kernel configurations based on model design and hardware characteristics. NASFLAT \cite{akhauri2024latency} combines an effective neural architecture sampler with supplementary encodings and transfer learning to optimize few-shot latency prediction. Our work integrates BO-based sampling to improve prediction reliability and efficiency further.
\textbf{\textit{Probabilistic predictors}}  offer insights into the confidence of the predictions, which can be useful to make more informed decisions during NAS. Among all the SOTA latency predictors accelerating NAS, our proposed predictor, HiFi-LLP, to the best of our knowledge, is the first to provide its predictions along with a confidence metric.

\section{Background}
% This section covers the fundamentals of our core components. We use Graph Neural Networks (GNNs) for DNN latency prediction by modeling DNN architectures as graphs, with nodes as layers and edges as dependencies. The message passing mechanism aggregates features from neighboring nodes to create a graph embedding, followed by a regression task to predict latency from the embedding.
%This section covers the fundamentals of our core components. 
We use Graph Neural Networks (GNNs) for DNN latency prediction by modeling DNN architectures as graphs with nodes as DNN layers, and edges as dependencies. The message passing mechanism aggregates features from neighboring nodes to create a graph embedding, followed by a regression task to predict the latency. % saved one line

\subsection{Graph Attention Networks (GAT)}
Among various GNN types, \textit{Graph Attention Networks (GAT)} enhance node representation learning using attention mechanisms. GATs address limitations of 
Graph Convolutional Networks (GCNs) by assigning different weights to neighboring nodes, allowing more context-aware information aggregation. GATv2 \cite{brody2021attentive}
% \textit{Graph Attention Networks v2 (GATv2)} 
employs a dynamic attention mechanism and is more expressive than GAT for capturing node interactions.

\subsection{Gaussian Process Regression}

Gaussian process (GP) regression is a non-parametric method for predicting uncertain quantities by modeling the distribution over possible functions that fit the observed data. %The essential concepts for this work are inducing points, kernel function, and output distribution.
Traditional GP regression has high computational complexity, making it impractical for large datasets. To solve this, approximated GP methods use \textit{inducing points}, i.e., selected representative points from the dataset. These points reduce the computational burden while maintaining the accuracy of the GP model. The inducing point positions are/can be optimized during training to represent the data well. \textit{Kernel Function} in GP regression determines how data points are related, capturing the smoothness and structure of the underlying function. The \textit{Matérn kernel} is a popular choice due to its ability to model various degrees of smoothness. The GP regression layer outputs a multivariate normal distribution, providing both the predicted value (mean) and its confidence (variance). This \textit{confidence} metric is key in our work to guide reliable exploration and optimization.

%\textbf{\textit{Inducing Points}}
%Traditional GP regression has high computational complexity, making it impractical for large datasets. To solve this, approximated GP methods use inducing points—selected representative points from the dataset. These points reduce the computational burden while maintaining the accuracy of the GP model. The inducing point positions are/can be optimized during training to represent the data best.
%\textbf{\textit{Kernel Function}}
%The kernel function in GP regression determines how data points are related, capturing the smoothness and structure of the underlying function. The Matérn kernel is a popular choice due to its ability to model various degrees of smoothness.
%\textbf{\textit{Output Distribution}}
%The GP regression layer outputs a multivariate normal distribution, providing both the predicted value (mean) and its confidence (variance). This confidence is key in our proposed hybrid NAS setup to guide reliable exploration and optimization.

\subsection{Bayesian Optimization}

Bayesian optimization (BO) is a sample-efficient method for optimizing expensive and complex black-box functions. %Its core components include a surrogate model, an acquisition function, and the use of prediction confidence. 
BO uses a \textit{surrogate model}, typically a GP, to approximate the target function. The mean predictions and uncertainty estimates provided by GP are crucial for guiding the optimization process. \textit{Acquisition function} directs the sampling step % process
by balancing exploration (sampling areas with high uncertainty) and exploitation (sampling areas with high objective values). Common acquisition functions, like Expected Improvement (EI) and Upper Confidence Bound (UCB), leverage the GP's prediction confidence to determine the most promising samples. The variance from the GP model indicates prediction confidence. By understanding this uncertainty, BO strategically selects samples to either explore uncertain regions or exploit areas with high objective values. This efficient strategy ensures rapid convergence to optimal solutions with fewer evaluations.

%\textbf{\textit{Surrogate Model}} 
%BO uses a surrogate model, typically a GP, to approximate the target function. The mean predictions and uncertainty estimates provided by GP are crucial for guiding the optimization process.
%\textbf{\textit{Acquisition Function}}
%The acquisition function directs the sampling process by balancing exploration (sampling areas with high uncertainty) and exploitation (sampling areas with high objective values). Common acquisition functions, such as Expected Improvement (EI) and Upper Confidence Bound (UCB), leverage the GP's prediction confidence to determine the most promising samples.
%\textbf{\textit{Leveraging Prediction Confidence}}
%The variance from the GP model indicates prediction confidence. By understanding this uncertainty, BO strategically selects samples to either explore uncertain regions or exploit areas with high objective values. This efficient strategy ensures rapid convergence to optimal solutions with fewer evaluations.

% In our latency predictor, BO effectively uses the GP's mean and variance to decide which architectures to evaluate next, optimizing the process by balancing exploration and exploitation based on prediction confidence.

% In our latency predictor setup, BO uses the GP model's mean and variance to make informed decisions about which architectures to evaluate next. This approach harnesses the GP's prediction confidence to effectively balance exploration and exploitation, optimizing the latency prediction process in a sample-efficient manner.

\section{HiFi-LLP Design}
\begin{figure}[h!]
  \centering
  \includegraphics[width=\columnwidth]{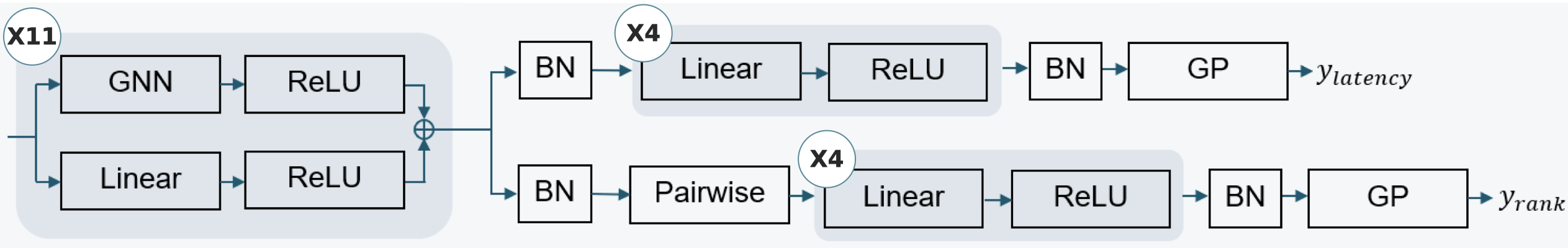}
  \caption{Architecture of HiFi-LLP. %The proposed predictor consists of a GNN backbone, whose output is passed through batch normalization layers and then duplicated for use in two separate heads. Both the latency predictor head and the rank predictor head include a \textit{FeatureExtractor} block, followed by a batch normalization layer and a GP head.
  }
  \label{fig:our_predcitor_architecture}
\end{figure}

HiFi-LLP is implemented using the PyTorch Geometric (PyG) library 
% \cite{pytorchGeometric} 
and is built upon GATv2 \cite{brody2021attentive}, incorporating 11 sequential layers with residual connections integrated with fully connected (FC) layers to improve training stability. 
Similar to HiFi-SAGE \cite{shambhavi25hifisage}, the model operates on batches of graphs, each representing a neural architecture. For a batch of size \(B\), the input consists of a node feature matrix of shape \([N, 59]\), an edge index matrix of shape \([2, E]\), and a batch vector of length \(N\). The node feature matrix encodes layer-specific attributes for each node in the architecture graph. The edge index matrix defines the graph connectivity, where each column specifies a directed edge between two nodes. The batch vector assigns each node to its corresponding architecture graph within the batch. To obtain a fixed-size representation of each architecture, node embeddings are aggregated using LCM readout \cite{ong2022learnable}, resulting in graph-level embeddings of shape \([B, 128]\). As illustrated in Fig.~\ref{fig:our_predcitor_architecture}, HiFi-LLP includes two prediction heads: a latency predictor and a pairwise rank predictor. The latency predictor takes the graph embedding and passes it through a batch normalization (BN) layer, followed by four linear FC layers and a GP regression head. This head produces both the predicted latency (mean) and its associated uncertainty (variance), enabling confidence-aware predictions. The pairwise rank predictor formulates latency comparison as a binary classification task. During training, all possible pairs of graph embeddings within a batch are considered. Each pair is concatenated, normalized, and passed through a stack of four FC layers, followed by a GP head. The output is a ranking score predicting which architecture is faster.

\subsection{Evaluation Metrics}
% \paragraph{Mean Absolute Percentage Error (MAPE)} It measures the average absolute deviation between predicted and actual values, expressed as a percentage. For $n$ samples, $A_t$ as the actual value, and predicted value $P_t$, MAPE is defined as:
% \begin{equation}\small
% \small  MAPE = \frac{1}{n}\sum_{t=1}^{n} \left| \frac{A_t - P_t}{A_t} \right| \times 100
% \end{equation}

% \vspace{-5mm} 

\paragraph{\textbf{Accuracy Bounds}} They represent the proportion of predictions with errors within thresholds (e.g., 10\%) of actual values. For \( n \) samples, let \( A_t \) be the actual value for observation \( t \), \( d_t \) the absolute difference between actual and predicted values, and \( \mathbf{1}[\cdot] \) an indicator function returning 1 if the condition is true, 0 otherwise. The error bound for an \(\alpha\%\) threshold is:
$\text{Acc}_{\alpha\%} = \frac{1}{n} \sum_{t=1}^{n} \mathbf{1}[d_t \leq \alpha \cdot A_t] \times 100$

% $\text{Acc}_{\alpha\%} = \frac{1}{n} \sum_{t=1}^{n} X_t \times 100$, where $X_t = 1$ if $d_t \leq \alpha \cdot A_t$, and $0$ otherwise.

% \begin{equation}\small
% Acc_{\alpha\%} = \frac{1}{n} \sum_{t=1}^{n} X_t \times 100 \quad \text{where} \quad X_t=
% \begin{cases}
%     1, & \text{if } d_t \leq \alpha \cdot A_t \\
%     0, & \text{otherwise}
% \end{cases}
% \end{equation}
% \vspace{-5mm}
\paragraph{\textbf{Spearman's Rank Correlation Coefficient (Fidelity score)}} It evaluates the monotonic relationship between predicted and actual rankings. Given $n$ samples and $d_t$ as the difference between the ranks of observation $t$, the fidelity score is defined as:
{$\rho = 1 - \frac{6\sum d_t^2}{n(n^2 - 1)}$}
% \end{equation}
% \vspace{-4mm}

\paragraph{\textbf{Entropy}}
We use Shannon's entropy to measure samples diversity. Higher entropy indicates more diverse samples, better capturing the dataset's structure. For a discrete set with probabilities $p(x)$, entropy is defined as:
$H(X)=-\sum_{x \in X} p(x) \log p(x) $

\subsection{Design Decisions}
\begin{figure}[h!]
  \centering
  \includegraphics[width=0.85\columnwidth]{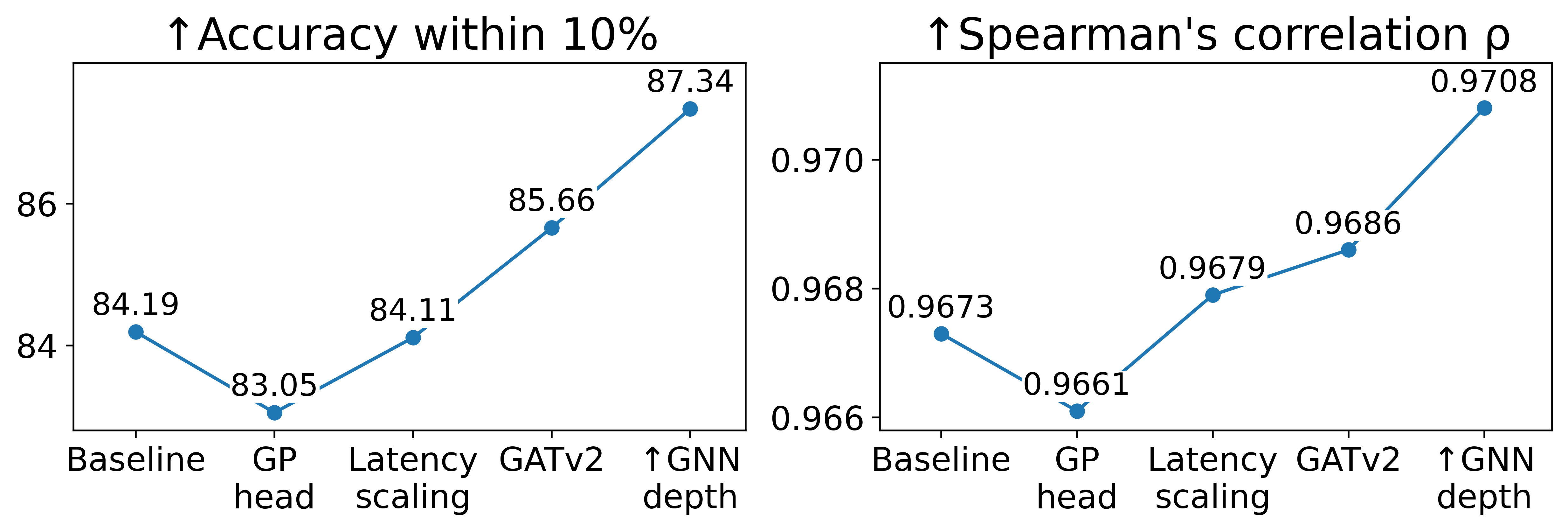}
  \caption{Ablation study of design decisions.
  %Starting from HiFi-SAGE, we incrementally modify the predictor.
  % by replacing the MLP head with a GP head, introducing latency scaling, switching GraphSAGE to GATv2, and increasing the GNN depth from 7 to 11 layers.
  }
  \label{fig:design_decisions_ablation}
\end{figure}

To evaluate the impact of key design choices, we started from HiFi-SAGE \cite{shambhavi25hifisage} as baseline and introduced incremental changes. Adding a GP head initially reduced performance due to its sensitivity to data scaling. To address this, we applied device-specific latency scaling by multiplying latency values with a scalar. In our experiments, this method was more effective than standard normalization (e.g., zero mean and unit variance). For each device, we trained the predictor with scalers in the range $[50,2000]$ (step size 50) and selected the one yielding the best performance which resulted in: \textit{DCPU: 400, DGPU: 250, MDSP: 800, EGPU: 100, ETPU: 750, MGPU: 100}. Thanks to HiFi-LLP's fast training time (under 15 minutes on a GTX 1080 Ti), this tuning was affordable in a few hours for all devices. The scaling restored performance,
% lowering MAPE 
and improved both the 10\% error bound and fidelity score. Finally, replacing the GraphSAGE backbone with GATv2 addressed the over-smoothing issue well known in GNNs and enabled a deeper network (11 layers), further boosting performance, as shown in Fig.~\ref{fig:design_decisions_ablation}.

\subsection{HW-aware Hybrid NAS framework}

\begin{figure}
    \centering
    \includegraphics[width=0.98\columnwidth]{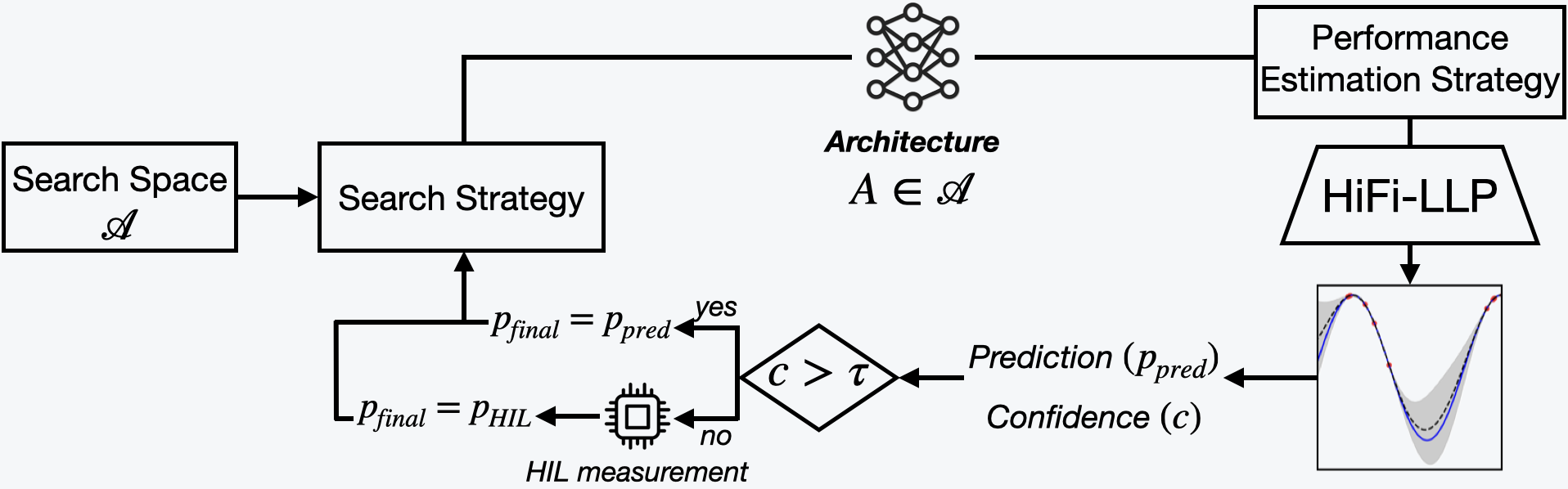}
    \caption{Our proposed hybrid NAS pipeline.}
    \label{fig:hybrid_nas_pipeline}
\end{figure}

Tackling the biggest challenge of reliability, especially in the sample-efficient training regime, HiFi-LLP makes it possible to design a hybrid NAS approach. As seen in Fig.  \ref{fig:hybrid_nas_pipeline}, for a given confidence threshold $\tau$, all the predictions falling below the threshold are re-evaluated on real HIL setup, while the confident predictions are retained for the search. For a poorly performing latency predictor, this would, in turn, end up becoming a HIL-guided search. Therefore, the trick lies in creating a predictor that offers a good trade-off between prediction accuracy, fidelity, and confidence metrics. Such a hybrid framework can directly leverage BO as a data sampler, allowing for the exploration of crucial regions of interest, owing to the confidence metric. 

\section{Evaluation}
\subsection{Experiment setups}

\begin{figure}[h!]
    \centering
    \includegraphics[width=\columnwidth]{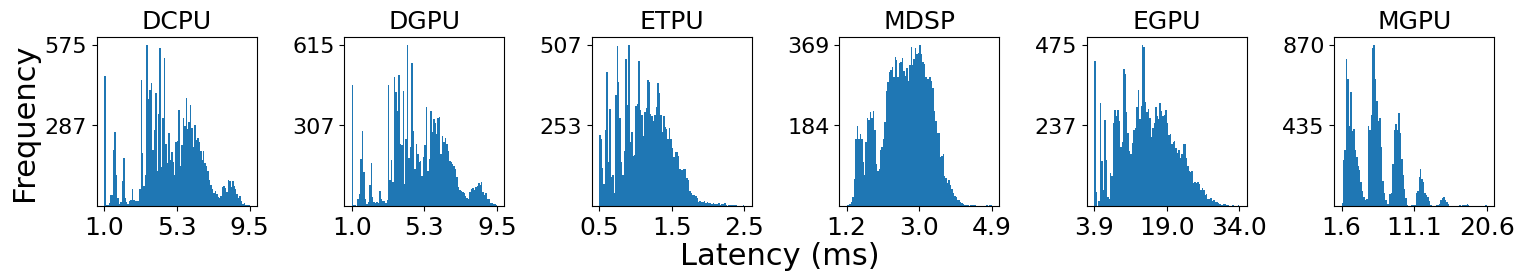}
    \caption{Latency distribution for devices in LatBench. %As can be seen, the existing devices have a broad range of latency distributions with ranging (e.g. compare ETPU and EGPU latency ranges). This highlights the importance of optimimzg the latancy scaler values indivudually for each device.
    }
    \label{fig:latency_distributions}
\end{figure}

Following \cite{dudziak2020brp} and \cite{shambhavi25hifisage}, we train and evaluate HiFi-LLP on LatBench \cite{dudziak2020brp}, an open-source benchmark containing precise latency measurements for NASBench-201 \cite{dong2020bench} architectures on different devices of varying scales (see Fig.~\ref{fig:latency_distributions}) namely, Intel Core i7-7820X (DCPU), NVIDIA GTX 1080 Ti (DGPU), NVIDIA Jetson Nano (EGPU), Google EdgeTPU (ETPU), Qualcomm Adreno 612 GPU (MGPU), and Qualcomm Hexagon 690 DSP (MDSP). Table \ref{tab:hyperparameters_optimal_values} details the optimal hyperparameter configurations and architectural choices.

\bgroup
\setlength{\tabcolsep}{3pt}
\begin{table}
    \centering
    \resizebox{\columnwidth}{!}{
    \renewcommand{\arraystretch}{0.85} % Reduce row height
    \begin{tabular}{l l | l l | l l}
        \toprule
        \textbf{Hyperparameter} & \textbf{Value} & \textbf{Hyperparameter} & \textbf{Value} & \textbf{Hyperparameter} & \textbf{Value} \\
        \midrule
        LR & $1e^{-3}$ & GNN type& GATv2& GP kernel& Matérn ($\nu$=5/2)\\
        LR schedule & Reduce on plateau & GNN aggregator& LCM& GP inducing points& 500 (learned)\\
        Optimizer & Adam & Aggregation size& 128& AE evolution cycles& 1000\\
        Batch size & 8 & GNN layers& 11& AE population size& 256\\
        Training epochs & 300 & GNN hidden size& 96& AE samples size& 128\\
        Early stopping & \cm (patience=30) & FC layers& 4& Acquisition function& UCB ($\beta$=0.5)\\
        Dropout (pairwise)& 0.01& FC hidden size& 64& \\
        \bottomrule
    \end{tabular}
    }
    \caption{Architectural choices and hyperparameter values.}
    \vspace{-0.1in}
    \label{tab:hyperparameters_optimal_values}
\end{table}
\egroup

\subsection{Accuracy Improvements} 
We assess the predictor's accuracy across multiple devices in extremely low (100 samples) and low (500 samples) data regimes. As shown in Table \ref{tab:accuracy_results_combined}, our model consistently outperforms the baselines. With 100 samples, it surpasses state-of-the-art HiFi-SAGE on 5 out of 6 devices and outperforms BRP-NAS on all devices. With 500 samples, it outperforms HiFi-SAGE across all devices with improvements in the 10\% error bound reaching up to 8.65 and 9.94 percentage points (p.p.), respectively. 
Our predictor performed the worst on the MGPU device when trained on 100 samples. We hypothesize that early stopping may have limited the model's ability to learn effectively. In our own experiments with a higher patience threshold, performance improved to 68\% for the 10\% error bound. However, to maintain consistency, we kept the setting the same and applied the same training procedure across all devices. Additionally, across all cases, our predictor showed significantly lower standard deviation in 10\% error bound (obtained from results of 3 different random seeds) than HiFi-SAGE—up to 13.62× and 10.53× lower in the 100- and 500-sample settings, respectively. Compared to BRP-NAS, it had a lower standard deviation on 4 of 6 devices, with a maximum reduction of 28.62×. This highlights HiFi-LLP's greater robustness and generalizability, which makes it less sensitive to specific data splits. Furthermore, averaging the accuracy across all six devices, our predictor consistently outperformed the baselines, achieving a 5.23 p.p. improvement with 500 samples and a 2.26 p.p. improvement with 100 samples over the SOTA.
\bgroup
\setlength{\tabcolsep}{2.5pt}
\begin{table*}
    \centering
    \vspace{0.11in}
    \resizebox{\linewidth}{!}{
    \begin{tabular}{l|ccc|ccc|ccc|ccc|ccc|ccc|ccc}
    \toprule
    \multirow{2}{*}{\textbf{\makecell{Model\\\scriptsize(\#train samples)}}} &
    \multicolumn{3}{c|}{\textbf{DCPU}} &
    \multicolumn{3}{c|}{\textbf{DGPU}} &
    \multicolumn{3}{c|}{\textbf{MDSP}} &
    \multicolumn{3}{c|}{\textbf{EGPU}} &
    \multicolumn{3}{c|}{\textbf{MGPU}} &
    \multicolumn{3}{c|}{\textbf{ETPU}} &
    \multicolumn{3}{c}{\textbf{Average}} \\[-3pt] 
    \cmidrule(lr){2-4} \cmidrule(lr){5-7} \cmidrule(lr){8-10} \cmidrule(lr){11-13} \cmidrule(lr){14-16} \cmidrule(lr){17-19} \cmidrule(lr){20-22}

    & $\uparrow$\textbf{\textit{10\%$_{Acc}$}} & $\uparrow$\textbf{\textit{5\%$_{Acc}$}} & $\uparrow$\textbf{\textit{1\%$_{Acc}$}} & 
        $\uparrow$\textbf{\textit{10\%$_{Acc}$}} & $\uparrow$\textbf{\textit{5\%$_{Acc}$}} & $\uparrow$\textbf{\textit{1\%$_{Acc}$}} &
        $\uparrow$\textbf{\textit{10\%$_{Acc}$}} & $\uparrow$\textbf{\textit{5\%$_{Acc}$}} & $\uparrow$\textbf{\textit{1\%$_{Acc}$}} &
        $\uparrow$\textbf{\textit{10\%$_{Acc}$}} & $\uparrow$\textbf{\textit{5\%$_{Acc}$}} & $\uparrow$\textbf{\textit{1\%$_{Acc}$}} &
        $\uparrow$\textbf{\textit{10\%$_{Acc}$}} & $\uparrow$\textbf{\textit{5\%$_{Acc}$}} & $\uparrow$\textbf{\textit{1\%$_{Acc}$}} &
        $\uparrow$\textbf{\textit{10\%$_{Acc}$}} & $\uparrow$\textbf{\textit{5\%$_{Acc}$}} & $\uparrow$\textbf{\textit{1\%$_{Acc}$}} &
        $\uparrow$\textbf{\textit{10\%$_{Acc}$}} & $\uparrow$\textbf{\textit{5\%$_{Acc}$}} & $\uparrow$\textbf{\textit{1\%$_{Acc}$}} \\[-3pt] 
    \midrule

    BRP-NAS\cite{dudziak2020brp}\scriptsize(\textbf{500}) & - & - & - & - & - & - & - & - & - & - & - & - & - & - & - & - & - & - & - & - & - \\
    HiFi-SAGE\cite{shambhavi25hifisage}\scriptsize(\textbf{500}) & $81.22_{2.6}$ & $53.94_{1.2}$ & $12.22_{0.3}$ & $85.56_{1.0}$ & $60.56_{3.2}$ & $13.95_{1.3}$ & $86.60_{2.1}$ & $59.61_{5.5}$ & $13.18_{2.5}$ & $85.51_{2.7}$ & $63.42_{4.2}$ & $15.08_{1.6}$ & $78.69_{4.1}$ & $\textbf{58.27}_{3.3}$ & $\textbf{14.44}_{1.1}$ & $80.04_{0.4}$ & $51.49_{1.5}$ & $11.49_{1.0}$ & $82.94$ & $57.88$ & $13.39$ \\
    Ours\scriptsize(\textbf{500}) & $\textbf{91.16}_{0.4}$ & $\textbf{72.33}_{0.7}$ & $\textbf{19.45}_{0.9}$ & $\textbf{92.05}_{0.9}$ & $\textbf{73.42}_{1.5}$ & $\textbf{20.23}_{0.8}$ & $\textbf{90.29}_{0.7}$ & $\textbf{67.21}_{1.4}$ & $\textbf{16.19}_{0.4}$ & $\textbf{92.23}_{0.7}$ & $\textbf{70.47}_{1.1}$ & $\textbf{17.54}_{0.9}$ & $\textbf{80.01}_{0.4}$ & $55.15_{2.3}$ & $13.04_{0.7}$ & $\textbf{83.29}_{0.9}$ & $\textbf{53.99}_{0.8}$ & $\textbf{11.75}_{0.2}$ & $\textbf{88.17}$ & $\textbf{65.43}$ & $\textbf{16.37}$ \\

    \midrule

    BRP-NAS\cite{dudziak2020brp}\scriptsize(\textbf{100})\normalsize$^\star$ & $51.5_{8.3}$ & $27.9_{5.5}$ & $6.1_{1.7}$ & $52.9_{5.0}$ & $28.7_{3.6}$ & $5.9_{1.3}$ & $78.4_{3.6}$ & $48.0_{3.8}$ & $10.3_{1.1}$ & $71.8_{3.5}$ & $\textbf{44.6}_{4.0}$ & $\textbf{9.9}_{1.3}$ & $46.3_{4.1}$ & $24.9_{3.4}$ & $5.2_{0.9}$ & $54.6_{5.7}$ & $30.0_{3.6}$ & $6.2_{1.0}$ & $59.25$ & $34.02$ & $7.27$ \\
    HiFi-SAGE\cite{shambhavi25hifisage}\scriptsize(\textbf{100}) & $61.65_{4.0}$ & $35.81_{4.3}$ & $7.66_{1.1}$ & $54.07_{3.7}$ & $28.71_{2.2}$ & $5.51_{0.5}$ & $74.96_{3.5}$ & $47.73_{2.6}$ & $10.21_{0.3}$ & $68.70_{8.5}$ & $40.89_{9.8}$ & $8.79_{2.5}$ & $\textbf{61.77}_{8.4}$ & $\textbf{37.60}_{6.7}$ & $\textbf{8.06}_{1.5}$ & $66.47_{3.7}$ & $38.32_{3.2}$ & $8.00_{0.8}$ & $64.60$ & $38.18$ & $8.04$ \\
    Ours\scriptsize(\textbf{100}) & $\textbf{66.15}_{2.3}$ & $\textbf{38.81}_{0.1}$ & $\textbf{8.21}_{0.3}$ & $\textbf{62.72}_{1.7}$ & $\textbf{36.48}_{1.8}$ & $\textbf{7.65}_{0.5}$ & $\textbf{82.63}_{0.3}$ & $\textbf{54.49}_{1.8}$ & $\textbf{12.17}_{0.7}$ & $\textbf{72.08}_{2.2}$ & $43.92_{1.9}$ & $9.31_{0.6}$ & $48.95_{5.3}$ & $28.88_{5.3}$ & $6.32_{1.5}$ & $\textbf{68.66}_{2.8}$ & $\textbf{39.66}_{2.7}$ & $\textbf{8.39}_{1.0}$ & $\textbf{66.86}$ & $\textbf{40.37}$ & $\textbf{8.67}$ \\

    \bottomrule
    \end{tabular}
    }
    \caption{Accuracy bound comparison across six hardware platforms and overall average. Reported values correspond to the mean and standard deviation for three random seeds. Models with $^\star$ use values reported from the corresponding paper.}
    % \caption{Accuracy bound comparison ($\uparrow$Acc$_{x\%}$) across six hardware platforms and overall average. Reported values correspond to the mean and standard deviation (rounded to 0.1) for models trained with 100 and 500 samples. Models with $^\star$ use values reported from prior work.}
    \label{tab:accuracy_results_combined}
\end{table*}
\egroup

%\input{data/tables/SOTA_eval_avg}

% Fig.~\ref{fig:prediction_vs_measurement} compares measured versus predicted latencies for our predictor and HiFi-SAGE, both trained on 500 samples, averaged over three runs for DCPU, DGPU, and EGPU. Our predictor consistently achieved lower MAPE, with up to 2.94 p.p. improvement. On DCPU and DGPU, gains are especially clear for latencies under 5 ms, and for EGPU, more predictions fall within the ±10\% error bound.

%% Behzad removed this
% Table \ref{tab:prediction_confidence_std} presents the average prediction STD. The GP regressor outputs both the prediction and confidence (STD) for each test sample in a single feedforward pass, which we leverage in our Hybrid NAS experiments.

% \begin{figure}
%     \centering
%     \includegraphics[width=.8\columnwidth]{data/figures/prediction_vs_measurement.png}
%     \caption{Prediction versus measured values. %The predicted values from our predictor and HiFi-SAGE, both trained on 500 samples, are plotted against the measured values. %Our predictor achieved a lower MAPE. The improvement is evident for DCPU in architectures with measured latencies below 5 ms, and for EGPU, where more predictions from HiFi-SAGE fall outside the ±10\% error bound compared to ours.
%     }
%     \label{fig:prediction_vs_measurement}
% \end{figure}

\subsection{Prediction Fidelity}
As an alternative to estimating absolute latency values, a reliable predictor may emphasize the relative ranking of input architectures based on their latencies \cite{dudziak2020brp}. Inspired by BRP-NAS \cite{dudziak2020brp}, we assess model predictions on the test set using 10 random seeds, average the rankings of each architecture, and compare them to ground truth rankings to report Spearman's rank correlation coefficient ($\rho$). 
We assess HiFi-LLP's performance against HELP \cite{lee2021help}, which is meta-trained and adapted to each device using 10 samples; BRP-NAS, trained with 900 samples per device; and HiFi-SAGE, trained with 100 samples per device. As shown in Table \ref{tab:prediction_fidelity}, HiFi-LLP enhances the mean Spearman's correlation across the three devices (DCPU, MDSP, and MGPU) by 1.85\% compared to BRP-NAS, and by 2.07\% over HELP. Furthermore, it surpasses HiFi-SAGE across all six devices with a mean improvement of over 0.2\%. Achieving an average correlation of 0.991, we demonstrate a strong fidelity in ranking architectures for latency.

\bgroup
\setlength{\tabcolsep}{3pt}
\begin{table}
    \centering
    \resizebox{0.9\columnwidth}{!}{
    \begin{tabular}{l|c|cccccc|cc} \toprule
        \multirow{2}{*}{\textbf{Model}} & \multirow{2}{*}{\textbf{\makecell{Target \\samples}}} & \multicolumn{6}{c}{$\uparrow$\textbf{Spearman's $\bm \rho$}} & \multirow{2}{*}{\textbf{Avg(3)}} & \multirow{2}{*}{\textbf{Avg(6)}}\\[-2pt] \cmidrule(lr){3-8}
         & & \textbf{\textit{DCPU}} & \textbf{\textit{DGPU}} & \textbf{\textit{MDSP}} & \textbf{\textit{EGPU}} & \textbf{\textit{MGPU}} & \textbf{\textit{ETPU}} \\ \midrule%\hline
         HELP$^\star$\cite{lee2021help} & 10 & 0.990 & - & 0.958 & - & 0.956 & - & 0.968 & - \\
         BRP-NAS$^\star$\cite{dudziak2020brp} & 900 & 0.991 & - & 0.959 & - & 0.961 & - & 0.970 & - \\
         HiFi-SAGE\cite{shambhavi25hifisage} & 100 & 0.991 & 0.991 & 0.985 & 0.995 & 0.981 & 0.992 & 0.986 & 0.989 \\
         Ours & 100 & \textbf{0.993} & \textbf{0.994} & \textbf{0.986} & \textbf{0.996} & \textbf{0.985} & \textbf{0.993} & \textbf{0.988} & \textbf{0.991} \\ 
         % Ours(100) & 100 & 0.994 & 0.994 & 0.986 & 0.996 & 0.986 & 0.993 & 0.989 & 0.991\\
         \bottomrule
    \end{tabular}
    }
    \caption{Predictions fidelity comparison. 
    %Spearman's rank correlation was computed between the predicted and ground truth rankings. Our predictor achieved the highest correlation values across all individual devices and in overall averages. \textbf{Mean(3)} represents the average correlation for DCPU, MDSP, and MGPU devices, while \textbf{Mean(6)} denotes the average across all six devices. 
    Results for models marked with $^\star$ are taken from their respective papers.}
    \label{tab:prediction_fidelity}
\end{table}
\egroup

% Table \ref{tab:prediction_fidelity} presents the Spearmans correlation values for different predictors. Our model outperforms BRP-NAS, which was trained on \textit{900 samples} per device, by improving the mean Spearman's correlation across the three devices reported by BRP-NAS (DCPU, MDSP, and MGPU) by 2.07\%. Furthermore, our predictor surpasses the state-of-the-art HiFi-SAGE on all six devices, achieving a mean improvement of more than 0.2\%. With an average correlation above \textit{0.99}, our predictor demonstrates strong fidelity in ranking architectures based on their latency.

\begin{figure}[h!]
    \centering
    \includegraphics[width=.98\columnwidth]{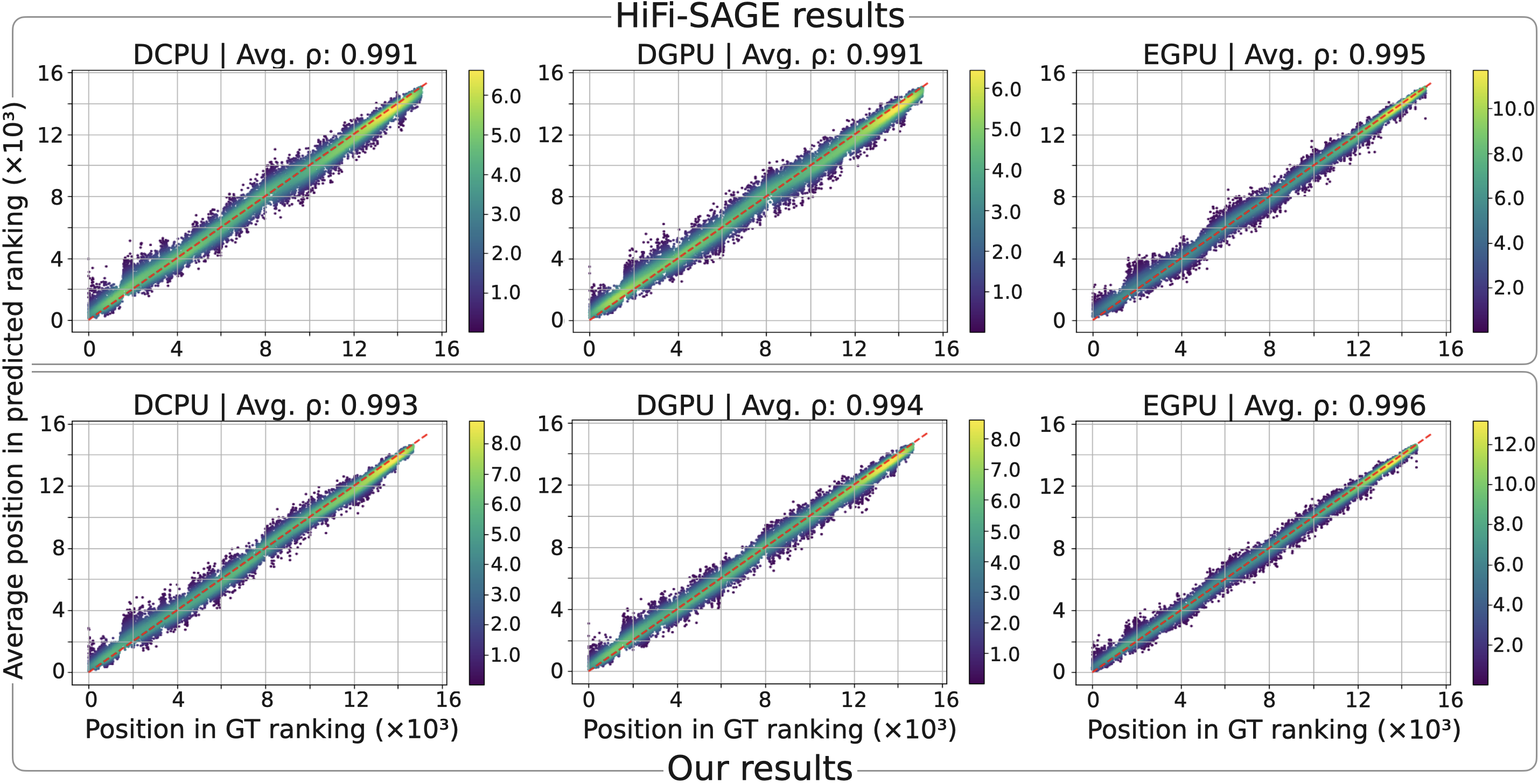}
    \caption{Density of correct latency ranking.}
    \vspace{-0.2in}
    \label{fig:prediction_fidelity}
\end{figure}

Fig.~\ref{fig:prediction_fidelity} shows the density of correct latency rankings, averaged over 10 runs, for our model and HiFi-SAGE, trained on 100 samples for the DCPU, DGPU, and EGPU devices. Our predictor maintains a more uniform density around the perfect ranking (diagonal line). For instance, for the EGPU device, HiFi-LLP has a higher density near the diagonal for lower-ranked architectures (positions 0 to 4). Additionally, our predictor's density distribution reaches higher upper bounds (12 and 8 vs. 10 and 6 for HiFi-SAGE), suggesting that even where HiFi-SAGE performed well, it predicted relatively fewer correct rankings.

\subsection{Oracle NAS}
\begin{figure}[t]
\centering
% \begin{subfigure}[t]{\textwidth}
%    \centering
%    \includegraphics[width=.95\linewidth]{data/figures/oracle_results.png}
%    \caption{Missed test accuracy and number of false positives/negatives.}
%    \label{fig:Ng1} 
% \end{subfigure}
\begin{subfigure}[t]{\linewidth}
   \centering
   \includegraphics[width=\linewidth]{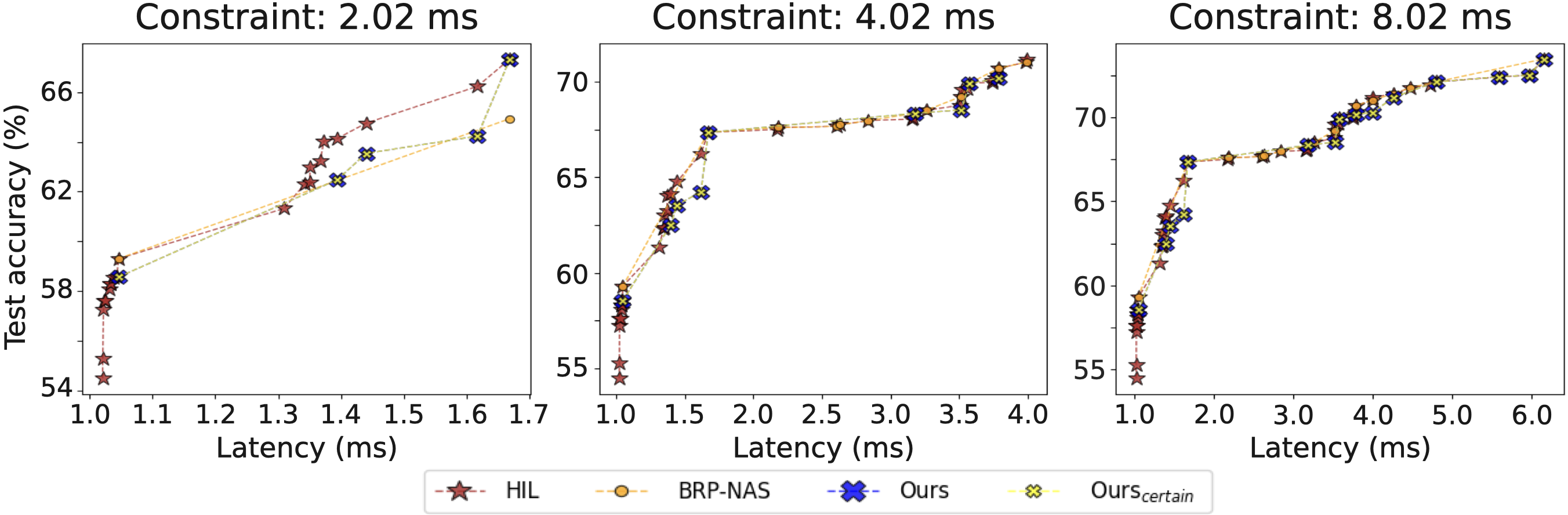}
   % \caption{}
   \label{fig:Ng2}
\end{subfigure}
\caption{Pareto front of Ours, BRP-NAS and HIL experiment.}
\vspace{-0.1in}
\label{fig:oracle_NAS}
\end{figure}

Oracle NAS \cite{dudziak2020brp} presumes perfect knowledge of architectures' accuracies and uses a latency predictor to estimate their latencies. It selects the most accurate architecture within the predicted latency constraint. A verification step checks the actual latency; if it surpasses the limit, the architecture is discarded. This process repeats with the next best candidate until a valid architecture is found. In Oracle NAS setting, we evaluate HiFi-LLP against BRP-NAS, trained on 200 samples from the DCPU device, using three metrics: missed test accuracy (the gap between selected and best achievable accuracy under the latency constraint), false positives (FP, models predicted to meet but fail the latency constraint), and false negatives (FN, models wrongly discarded based on predicted latency). Experiments spanned latency constraints from 1.12 to 9.32, with a step size of 0.1. To enhance reliability, we compare a confidence-aware variant of our predictor, named \textit{Ours$_{certain}$} in Table \ref{tab:oracle_NAS_results}, designed to filter out predictions with high uncertainty. Specifically, we exclude any architecture where the standard deviation (STD) of the predicted latency surpasses 1.1778. This threshold, representing the 90th percentile of HiFi-LLP's prediction STDs across the DCPU dataset, acts as a practical heuristic. Table \ref{tab:oracle_NAS_results} shows the average missed test accuracy, FP, and FN. Initially, our predictor had a lower missed test accuracy compared to BRP-NAS but exhibited a higher number of false positives. However, \textit{Ours$_{certain}$} reduced FP and FN by 2.04\% and 23.3\%, respectively. Under different latency constraints, Fig.  \ref{fig:oracle_NAS} displays the Pareto optimal solutions for our predictors and BRP-NAS, along with the Pareto front obtained using HIL measurements. Our predictor's Pareto optimal solutions closely matches the HIL Pareto front, suggesting that HIL measurements could be replaced by our predictor with similar results.

\begin{table}
    \centering
    \resizebox{0.7\columnwidth}{!}{
    \begin{tabular}{l|ccc} \toprule
       \textbf{Model}  &  \makecell{\textbf{↓Missed test} \\\textbf{accuracy(p.p.)}} & \textbf{↓FP} & \textbf{↓FN}\\ \midrule%\hline
        BRP-NAS\cite{dudziak2020brp} & 0.42 & 229.93 & 284.22\\
        Ours & 0.26 & 255.96 & 234.76\\
        Ours$_{certain}$ & \textbf{0.26} & \textbf{225.23} & \textbf{217.98}\\
        \bottomrule
    \end{tabular}
    }
    \caption{Comparison of latency predictors in Oracle NAS. 
    %The average missed test accuracy, false positives (FP), and false negatives (FN) are reported. Initially, our predictor had a lower missed test accuracy compared to BRP-NAS but showed a higher number of false positives. However, Ours$_{certain}$ was able to reduce the number of false positives and false negatives by 2.04\% and 23.30\%, respectively.
    }
    \label{tab:oracle_NAS_results}
\end{table}

% \begin{figure*}[t]
% \centering
% \begin{subfigure}[t]{.48\textwidth}
%    \centering
%    \includegraphics[width=.95\linewidth]{data/figures/oracle_results.png}
%    \caption{Missed test accuracy and number of false positives/negatives.}
%    \label{fig:Ng1} 
% \end{subfigure}
% \begin{subfigure}[t]{.48\textwidth}
%    \centering
%    \includegraphics[width=.95\linewidth]{data/figures/oracle_pareto.png}
%    \caption{Pareto optimals for predictors and Pareto frontier using HIL.}
%    \label{fig:Ng2}
% \end{subfigure}

% \caption{Comparison of predictors in an Oracle NAS setup.}
% \label{fig:oracle_NAS}
% \end{figure*}

\subsection{Hybrid NAS}
We assess the reliability of our hybrid NAS against standard NAS under identical conditions, differing solely in prediction usage: standard NAS consistently relies on the predictor's output, whereas our hybrid NAS discards predictions with uncertainty (STD > 1.1778) and defaults to HIL measurements. We compare the relative runtime of BRP-NAS and our predictor in both configurations. On the NVIDIA GeForce GTX 1080 Ti, BRP-NAS and HiFi-LLP require 0.001 and 0.21 seconds per inference, respectively. We presume HIL measurements take 1 minute per compilation, transfer, and inference, similar to \cite{shambhavi25hifisage}. All predictors were trained using 200 samples from the DCPU device. Table \ref{tab:hybrid_NAS_resultss} presents NAS results using the Aging Evolution (AE) \cite{real2019regularized} search algorithm, where predictors selected architectures \textit{assumed} to be optimal, without verifying against ground-truth measurements. BRP-NAS failed to meet the predefined latency constraint in all cases, while our predictor succeeded in 2/3 cases. In the hybrid setup, our predictor not only consistently met the constraint but also achieved up to an 8.6x speedup compared to full HIL-based NAS.
\bgroup
\setlength{\tabcolsep}{4pt}
\begin{table}
    \centering
    \vspace{0.11in}
    \resizebox{0.85\columnwidth}{!}{
    \begin{tabular}{l|c|ccc|cc|c} \toprule
        \textbf{Model} & \makecell{\textbf{Const.} \\[-2pt] \textbf{(ms)}} & \makecell{\textbf{Lat.} \\[-2pt] \textbf{(ms)}} & \makecell{\textbf{Acc.} \\[-2pt] (\textbf{\%})} & \makecell{\textbf{Is} \\[-2pt] \textbf{valid}} & \makecell{\textbf{Evaluated} \\[-2pt] \textbf{models}} & \makecell{\textbf{HIL} \\[-2pt] \textbf{runs}} & \makecell{\textbf{Relative Avg.} \\[-2pt] \textbf{run time}} \\ \midrule%\hline

        % HIL & \multirow{4}{*}{2.02} & 1.66 & 66.36 & \cm & 5665 & 5665 & 1 & 1.0x\\
        BRP-NAS\cite{dudziak2020brp} & \multirow{3}{*}{2.02} & 2.12 & 65.92 & \xm & 5220 & - & 0.00001\\
        Ours & & 3.18 & 67.50 & \xm & 9958 & - & 0.003\\
        Ours$_{hybrid}$ & & 1.72 & 64.94 & \cm & 6556 & 1439 & 0.2222\\ \midrule%\hline

        % HIL & \multirow{4}{*}{4.02} & 3.99 & 71.18 & \cm & 2114 & 2114 & 1 & 1.0x\\
        BRP-NAS\cite{dudziak2020brp} & \multirow{3}{*}{4.02} & 4.97 & 71.22 & \xm & 2183 & - & 0.00001\\
        Ours & & 3.82 & 70.22 & \cm & 2680 & - & 0.003 \\
        Ours$_{hybrid}$ & & 3.82 & 70.22 & \cm & 2680 & 301 & 0.1154\\ \midrule%\hline

        % HIL & \multirow{4}{*}{8.02} & 6.15 & 73.02 & \cm & 1073 & 1073 & 1 & 1.0x\\
        BRP-NAS\cite{dudziak2020brp} & \multirow{3}{*}{8.02} & 8.39 & 72.72 & \xm & 1065 & - & 0.00001\\
        Ours & & 6.94 & 73.20 & \cm & 1097& - & 0.003\\
        Ours$_{hybrid}$ & & 6.15 & 73.02 & \cm & 1073 & 134 & 0.1279\\
        \bottomrule
    \end{tabular}
    }
    \caption{Performance comparison of latency predictors in a latency-constrained NAS setup. Abbreviations: \textbf{Const.} = Constraint, \textbf{Lat.} = Latency, \textbf{Acc.} = Accuracy, \textbf{Avg.} = Average.}
    %\vspace{-0.1in}
    % Unlike BRP-NAS, which consistently exceeded the latency constraint, our predictor met the constraint in two out of three cases. In a hybrid setup, our predictor consistently adhered to the constraint, achieving up to 8.6x speedup over HIL measurements.
    %While BRP-NAS could never adhere to the predefined latency constraint (i.e., the architecture it selected had a latency higher than the threshold), our predictor was able to perform better and meet the constraint in two out of three cases. Additionally, when we used our predictor in a hybrid setup, it always adhered to the constraint and achieved up to 8.6x speedup compared to HIL measurements.
    % }
    \label{tab:hybrid_NAS_resultss}
\end{table}
\egroup

\subsection{Bayesian Optimization (BO)-based Data Sampling}
\begin{figure}
  \centering
  \includegraphics[width=.8\columnwidth]{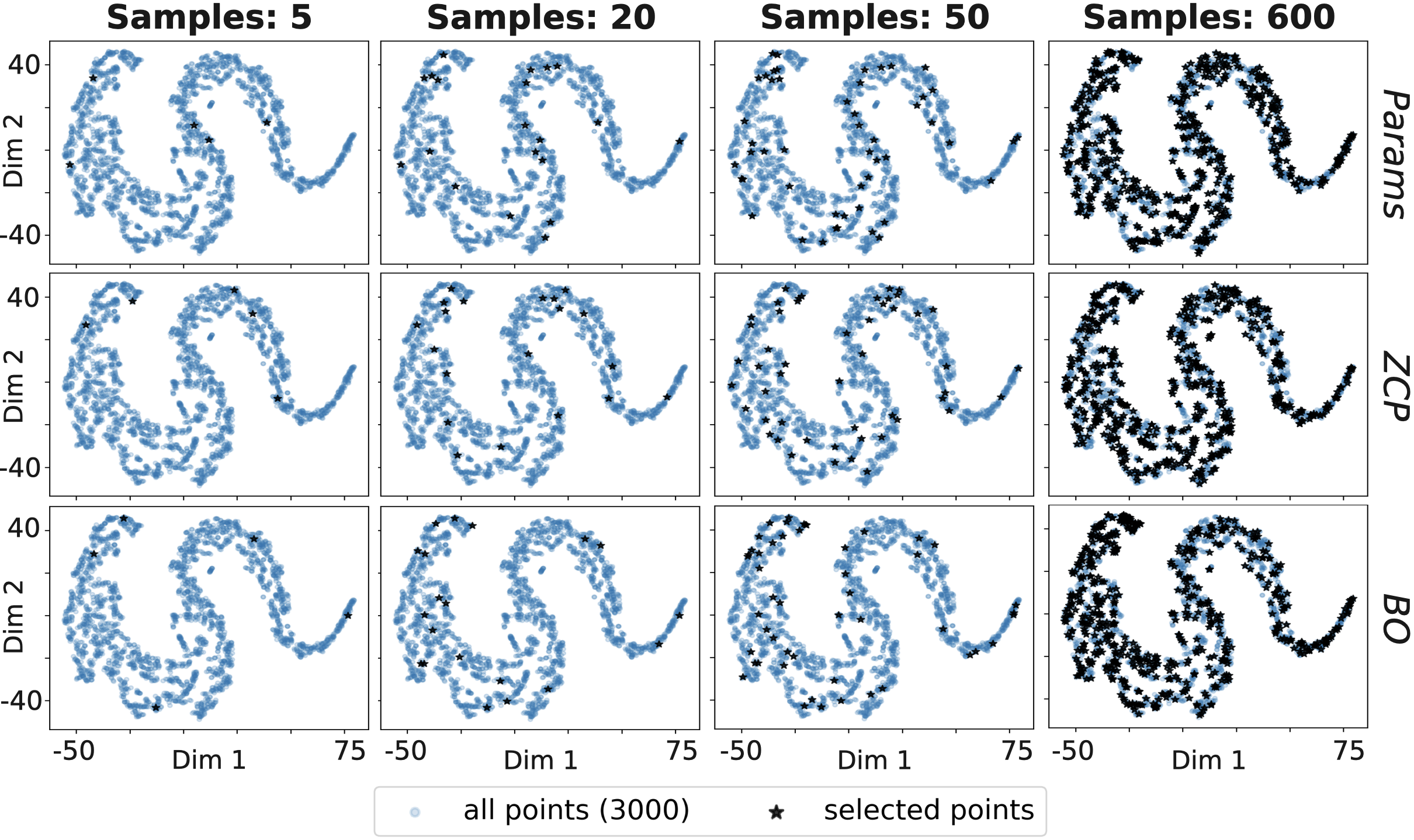}
  \caption{Qualitative comparison of various samplers.}
  \vspace{-0.2in}
  \label{fig:samplers_comparison}
\end{figure}

In this experiment, we simulate a real-world scenario where an existing predictor needs to adapt to previously unseen measurements, evaluating how different samplers influence this adaptation process. We use a BO objective to measure the percentage error between predicted and actual latencies. This is calculated for all samples in the adaptation pool. Since BO updates its model iteratively, selecting all (k) adaptation samples at once is not practical. We start with (k' = 10) initial samples, train the model, and iteratively add (k') samples based on updated predictions. After each iteration, the model retrains on all accumulated samples for 10 epochs. This method is consistently applied to all samplers for fairness, even if some can select (k) samples in one go. We adapt from a DCPU model (pretrained on 300 samples) to ETPU, using up to (k = 600) samples, and also report results for 300 and 400 samples. The sampling pool consists of 3000 architectures. We assess the 10\% accuracy bound, improvement over random sampling, and the entropy of selected samples.
Table \ref{tab:samplers_comparison} compares random sampling, SOTA stratified, encoding-based, and BO-based samplers at 300, 400, and 600 samples. In low-data scenarios, our sampler outperformed others, with accuracy gains of 0.325 and 4.519 p.p. at 300 and 400 samples, respectively, and an improved entropy. At 600 samples, test accuracies converged across samplers, with BO showing the highest entropy. These results emphasize the importance of sampler choice in low-data settings.
Fig.~\ref{fig:samplers_comparison} illustrates the 2D t-SNE \cite{van2008visualizing} reduction of architecture graph embeddings, highlighting samples chosen by each sampler. Notably, BO selects more informative and diverse samples early, such as at  $samples = 5$,  where it uniquely picks from the right tail, exploring subspaces overlooked by other samplers.

\bgroup
\setlength{\tabcolsep}{3pt}
\begin{table}[tbp]
    \centering
    \vspace{0.11in}
    \resizebox{0.85\columnwidth}{!}{
    \begin{tabular}{l|cc|cc|cc}
        \toprule \multicolumn{1}{l}{\multirow{2}{*}{\textbf{Sampler}}} & \multicolumn{2}{c}{\textbf{samples=300}} & \multicolumn{2}{c}{\textbf{samples=400}} & \multicolumn{2}{c}{\textbf{samples=600}}\\[-2pt]
       \cmidrule(lr){2-3} \cmidrule(lr){4-5} \cmidrule(lr){6-7} 
         & $\uparrow$\textbf{\textit{10\%$_{Acc}$}}  & $\uparrow$\textbf{\textit{Entr.}}  & $\uparrow$\textbf{\textit{10\%$_{Acc}$}} &  $\uparrow$\textbf{\textit{Entr.}}  & $\uparrow$\textbf{\textit{10\%$_{Acc}$}} & $\uparrow$\textbf{\textit{Entr.}}  \\ \midrule%\hline
        Random & 74.247  & 5.594 & 79.288  & 5.871 & 85.510  & 6.255 \\
        Accuracy & 78.389 & 5.602  & 80.118 &  5.873  & \textbf{89.969} &  6.255 \\
        Parameters & \underline{78.706} &  5.602  &  81.624 &  5.873 & 84.132 &  6.255  \\
        Latency & 73.117 &  \underline{5.603} & 78.192 &  5.868 & 88.112 &  6.257\\ %\hline
        \midrule
        Arch2Vec & 78.295 &  5.586 & 80.880 &  5.868  & 89.156 &  6.259 \\
        CATE & 77.063 &  5.578 & 80.888 &  5.846 & 85.664 &  6.238\\
        ZCP & 78.038 &  5.597 & \underline{81.684} &   \underline{5.874} & \underline{89.858} &  \underline{6.267}\\
        CAZ & 75.428 &  5.586 & 78.946 &  5.872 & 88.643 &  6.263\\ %\hline
        \midrule
        BO \textbf{(Ours)} & \textbf{79.031} &  \textbf{5.612} & \textbf{86.203} &  \textbf{5.885} & 89.841 &  \textbf{6.280}\\
        \bottomrule
    \end{tabular}
    }
    \caption{Comparison of different SOTA samplers for adapting a pretrained predictor from DCPU to ETPU dataset. SOTA samplers can be found in \cite{akhauri2024latency}.
    }
    \vspace{-0.2in}
    \label{tab:samplers_comparison}
\end{table}
\egroup

\section{Conclusion}
This paper presents HiFi-LLP, a novel platform-specific latency predictor that excels with minimal training data, requiring only 100 samples to deliver predictions with a confidence score. Our key innovation lies in the design of a GATv2-based network paired with a Gaussian process regression head, crafted for superior latency and rank predictions. HiFi-LLP stands out as a fast, cost-effective predictor while maintaining high reliability for NAS applications. By utilizing the confidence score, we have developed a robust hybrid HW-aware NAS framework. Our experiments demonstrate that Bayesian optimization-based data sampling surpasses other state-of-the-art methods in low-data environments. Looking ahead, future work will focus on integrating our hybrid NAS strategy with the Bayesian optimization sampler, aiming to optimize the exploration-exploitation balance in NAS methods, ultimately enhancing both the speed and quality of NAS applications.

\bibliographystyle{IEEEtran}
\bibliography{sources}
\end{document}